\documentclass{article}

\usepackage{microtype}
\usepackage{graphicx}
\usepackage{subcaption}
\usepackage{booktabs} 

\usepackage{hyperref}

\usepackage{tikz}
\usetikzlibrary{arrows.meta, positioning, shapes.geometric}

\usepackage[preprint]{icml2026}

\usepackage{amsmath}
\usepackage{amssymb}
\usepackage{mathtools}
\usepackage{amsthm}
\usepackage{enumitem}
\setlist{nosep}

\usepackage{bm}
\usepackage{bbm}
\usepackage{comment} 

\usepackage[capitalize,noabbrev]{cleveref}

\theoremstyle{plain}

\theoremstyle{definition}

\theoremstyle{remark}

\icmltitlerunning{Data Manipulation Attacks Against ML-Integrated CPS}

\begin{document}
\setlength{\abovedisplayskip}{3pt}
\setlength{\belowdisplayskip}{3pt}

\twocolumn[
  \icmltitle{Data Manipulation Attacks Against Machine Learning Integrated Cyber-Physical Systems}



  \icmlsetsymbol{equal}{*}

  \begin{icmlauthorlist}
    \icmlauthor{Soyon Choi}{vandy}
    \icmlauthor{Scott Alfeld}{amh}
    \icmlauthor{Meiyi Ma}{vandy}
  \end{icmlauthorlist}

  \icmlaffiliation{amh}{Department of Computer Science, Amherst College, Amherst, MA}
  \icmlaffiliation{vandy}{Department of Computer Science, Vanderbilt University, Nashville, TN}

  \icmlcorrespondingauthor{Soyon Choi}{soyon.choi@vanderbilt.edu}
  \icmlcorrespondingauthor{Scott Alfeld}{salfeld@amherst.edu}

  \icmlkeywords{Machine Learning, Adversarial Learning}

  \vskip 0.3in
]



\printAffiliationsAndNotice{}  

\begin{abstract}
There is a well-established body of work demonstrating that individual machine learning (ML) algorithms can be manipulated or deceived.
However, in many real-world systems, multiple ML algorithms are composed to enable automated decision-making, particularly in Cyber-Physical Systems (CPS).
Often such systems use a composition of vision, planning, and other ML models to affect their internal state and ultimate take actions in the physical world.
While there is a huge literature on the (in)security of ML algorithms, it is not immediately clear how an attacker might leverage their ability to trick individual learners when attacking the composition as a whole.
We develop methods for creating an {\it attack policy} (a mapping from state to attack) for a data-manipulating attacker affecting a CPS or other state-machine-based AI compositions.
We describe our framework, a basic threat model and corresponding methods for computing an optimal attack policy, as well as planned extensions.
The goal of this work-in-progress paper is to stimulate conversation and forge new collaborations regarding the security of compositional learning.
\end{abstract}

\section{Introduction}

Machine learning (ML)  methods, while powerful, are vulnerable to data manipulation attacks.
Attacks that manipulate input data (whether at training or deployment time) to change predictions are well-studied \cite{szegedy_2014_intriguing, alfeld_2016, fawzi_2018, machado_silva_2021}.
Many real-world systems, however, are a composition of ML methods and as such their vulnerabilities are less clear than those of simple learning algorithms.
In this work we focus on systems where the composition of learners is made explicit via a state machine.
We are motivated in particular by deployed Cyber-Physical Systems (CPS) which are often composed of multiple ML subsystems, as well as various input sources (e.g., sensors, cameras, user commands), controllers, and actuators \cite{muntean_2022, deshpande_2023, guo_2023}.
Similarly to the vulnerability of individual ML models, the vulnerability of specific components of a CPS to adversarial manipulation is well-documented, particularly regarding vision-based perception \cite{eykholt_2018}.
However, the components of a CPS interact in complex ways, resulting in intricate state transitions that ultimately dictate safety-critical decisions.

In this on-going work, we consider an attacker aiming to manipulate the behavior of CPS by exploiting vulnerabilities within its underlying ML model components.
We assume the attacker has limited ability to affect some or all of the individual ML components.
We propose a generalized framework by which an attacker can use this ability to make localized perturbations to achieve broader, system-level goals.
We also provide an open-source python implementation of our methods to the community.
Concretely, we make the following key contributions:
\begin{enumerate}
    \item We propose a framework for constructing adversarial policies that leverage localized manipulation of ML components to achieve system-level objectives against ML-integrated Cyber-Physical Systems.
   \item We show that optimal attack actions depend not only on the current state of the defender but on the other parameters of the state machine.
    \item We introduce CP-ATK, a software package for computing optimal attack policies given a state-machine description. 
\end{enumerate}

\section{Background}
\setlength{\belowcaptionskip}{-1.0cm}
\begin{figure}[t]
    \centering
    \resizebox{0.9\linewidth}{!}{
    \begin{tikzpicture}[
        node distance=1.2cm and 1.5cm,
        state/.style={
            rectangle,
            draw=black!70,
            fill=black!5,
            very thick,
            rounded corners=5pt,
            minimum width=2.5cm,
            minimum height=1cm,
            align=center,
            font=\sffamily\small
        },
        arrow/.style={-{Stealth[length=5pt]}, thick},
        label style/.style={
            font=\small,
            inner sep=2pt
        }
    ]
        \node[state] (stopped) {$\texttt{Stopped~(S)}$};
        \node[state, below=of stopped] (accel) {$\texttt{Accelerating~(A)}$};
        \node[state, below=of accel] (lane) {$\texttt{Lane Keeping~(LK)}$};
        \node[state, left=of lane, xshift=-0.5cm, yshift=-1.5cm] (deaccel) {$\texttt{Decelerating~(D)}$};
        \node[state, right=of lane, xshift=0.5cm, yshift=-1.5cm] (park) {$\texttt{Parking~(P)}$};
        \node[state, below=of lane] (pass) {$\texttt{Passing~(PS)}$};
    
        \path[-{Stealth[length=6pt]}, thick]
            (stopped) edge node[right] {$\alpha_{S,A}$} (accel)
            
            (accel) edge[bend left=20] node[right, pos=0.5] {$\alpha_{A,LK}$} (lane)
            (lane) edge[bend left=20] node[left, pos=0.5] {$\alpha_{LK,A}$} (accel)
            
            (lane) edge[bend right=20] node[above left] {$\alpha_{LK,D}$} (deaccel)
            (lane) edge[bend left=20] node[above right] {$\alpha_{LK,PK}$} (park)
            
            (lane) edge[bend left=20] node[right, pos=0.5] {$\alpha_{LK,PS}$} (pass)
            (pass) edge[bend left=20] node[left, pos=0.5] {$\alpha_{PS,LK}$} (lane)
            
            (deaccel) edge[bend left=35] node[left, pos=0.7] {$\alpha_{D,S}$} (stopped)
            (park) edge[bend right=35] node[right, pos=0.7] {$\alpha_{PK,S}$} (stopped);
    \end{tikzpicture}
    }
    \caption{{\bf Self Driving Car Example.} 
    The $\texttt{Stopped~(S)}$ state represents the car stopped, awaiting instruction. 
    The $\texttt{Accelerating~(A)}$ state represents a car that is increasing its velocity. 
    When moving at the desired speed in the desired lane with no perceived obstacles, the car is in the $\texttt{Lane Keeping~(LK)}$ state. 
    The $\texttt{Decelerating~(D)}$ and $\texttt{Passing~(PS)}$ states represent reactive maneuvers to environmental hazards, while $\texttt{Parking~(PK)}$ represents a transition to a terminal safety state. 
    In our example, the attacker’s objective is to reach the target state $s^\dagger = \texttt{S}$; the optimal adversarial policy shifts between exploiting the $\texttt{D}$ or $\texttt{PS}$ pathways depending on the defender's transition probabilities, defined by the parameters $\alpha_{s,s'}$ (transition probability from state $s$ to state $s'$).
    For figure clarity we have removed the self-loops of every state, representing the fact that the car may remain in a state across multiple time steps.
    }
    
    \label{fig:sdc}
\end{figure}
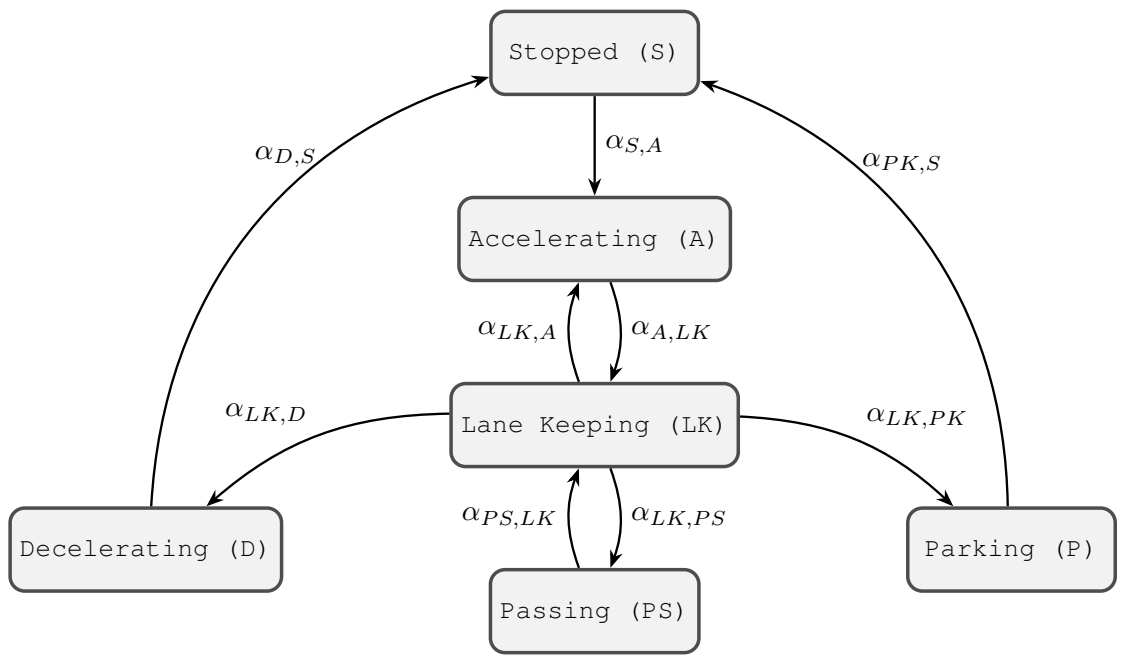

Many CPS are translatable into or explicitly modeled as Finite State Machines (FSMs) to manage the complexity of high-level decision-making and planning \cite{badue_2020, gonzalez_2022}.
In autonomous driving specifically, FSMs serve as the hierarchical bridge between low-level control (e.g., PID or MPC) and high-level route planning \cite{ghraizi_2023}.

We capture the FSM of any such CPS with the following notation.
We denote the set of states as \(\mathcal S\) and the set of all possible actions as \(\mathcal A\).
For each state \(s \in \mathcal S\) we denote the set of available actions in that particular state as \(\mathcal A(s)\).
Note that some actions change the underlying internal state (e.g., ``object approaching at high speed detected''), while others correspond to actions in the physical world (e.g., ``apply the breaks'' or ``change lane'').
Also in every state \(s\), we assume a collection of ML models \(\mathcal M(s)\) are used.
The CPS, at state \(s\), acts based on its decision functions \(F_s\) which take the outputs \(o\) of \(\mathcal M(s)\) and return an element of \(\mathcal A(s)\).
For example, a curb-detection model (among others) is used when a self-driving car is parking whereas a lane detection model is used on the highways.

\section{Threat Model}

Toward our ultimate aim of creating more robust CPS, we seek to develop a generalized framework by which optimal attack policies can be computed for a broad class of threat models.
In what follows, we characterize a subset of specific threat models for which we have developed tractable methods to compute such attack policies.

{\bf Attacker Knowledge.}
We consider a gray-box setting in the following sense.
We assume the attacker has full knowledge of the defender's underlying FSM.
However, the attack has only limited knowledge about the ML models used in each state.
This (lack of) knowledge about the individual models leads to the probabilistic capabilities described next.

{\bf Attacker Capability.}
For illustrative purposes, we consider a simple model and present a detailed, exact solution. 
We discuss more complicated models in Section~\ref{sec:extensions}.
We assume that in every state \(s \in \mathcal S\), the attacker has (known) probability \(\alpha_{s, a^\dagger}\) of being able to attack the ML models used in \(s\) so as to pick the resulting action \(a^\dagger \in \mathcal A(s)\).
An attack policy is then a mapping from \(\mathcal{S}\) to \(\mathcal{A}\) which determines the best action \(a^\dagger\) to select in each stae

In addition, we assume an attacker with a model of how the CPS behaves under normal inputs. 
Namely, the attacker has a distribution \(P(a | s)\) estimating the probability that the defender takes action \(a\) while in state \(s\) if no attack occurs.

{\bf Attacker Goal.}
We consider an attacker with the goal of causing the defender to enter a target state \(s^\dagger \in \mathcal S\).
For mathematical elegance we consider a soft horizon --- the attacker wants the defender to enter that state as soon as possible, rather by a hard deadline.
Alternative threat models, including with a hard deadline, are discussed in Section~\ref{sec:extensions}.

{\bf Example: Self Driving Car.}
As an illustrative example we consider the simplified model of a self driving car's underlying FSM \cite{ghraizi_2023}.
The model is pictured in Figure~\ref{fig:sdc} (see caption for explanation of individual states).
The agent has several sensors, including cameras, a GPS, engine telemetry, an emergency RF receiver, and LIDAR.

In each state \(s \in \mathcal S\), various ML models are used to determine the action taken, and how they do so (the \(F_s\) functions) differ across states.
For example, a curb-detection model (among others) is used in \texttt{Parking} whereas a lane detection model is used in \texttt{Lane Keeping}.
As some models and sensors are more vulnerable to data manipulation than others, every state's \(\alpha\) value is (potentially) unique.
Some actions (e.g., activating the turn signal) keep the agent in the same state, while others (e.g., transitioning from \texttt{Stopped} to \texttt{Accelerating}) change the internal state.

We use this state machine to illustrate a fundamental property of attack policies.
Namely, {\it the optimal attack action for any given depends on non-local probabilities.}

Consider an attacker with the target state \(s^\dagger = \texttt{Stopped}\).
The success of an attack depends largely on which sensors or behaviors the adversary targets. 
For example, tricking a camera into seeing a ``ghost'' obstacle is relatively simple, whereas spoofing emergency radio signals vehicles is much more difficult and higher risk for the attacker.
In what follows we consider several underlying CPS configurations and describe their effects on the optimal attack.

Consider first the simplest scenario where all transition probabilities are \(1 / |\mathcal{A}(s)|\) and all \(\alpha\)'s are\footnote{For notational convenience in this example we use \(\alpha_{s, s'}\) to denote \(\alpha_{s, a}\) where \(a\) is the action that moves from \(s\) to \(s'\).} equal except for \(\alpha_{D, S}\) and \(\alpha_{PK, S}\).
Because of the symmetry of the state machine, the optimal attack action for the state \texttt{LK} depends entirely on the relationship between \(\alpha_{D, S}\) and \(\alpha_{PK, S}\). 
If \(\alpha_{D, S}\) (resp. \(\alpha_{PK, S}\)) is smaller, then the attacker will favor moving the defender to \texttt{Decelerating} (resp. \texttt{Parking}).
Note the non-local effects: 
If state \(D\) uses inputs from engine telemetry and state \(P\) uses GPS data, then the vulnerability of the ML models used on such data determines the attack action for state \(LK\) even if \(LK\) ignores telemetry and GPS inputs.

In general, of course, transition probabilities are not all identical and the \(\alpha\) terms will differ due to the robustness of both sensors and models.
If the car is programmed to be highly sensitive to hazards, it will favor the \texttt{Decelerating} state over \texttt{Passing}. 
In this case, the attacker’s optimal policy is straightforward: trick the camera/vision model into perceiving a minor obstacle. 
Then, the car will follow a low-cost path from \texttt{Lane~Keeping} to \texttt{Decelerating}, and eventually to the target \texttt{Stopped} state.

If instead the car is tuned to maintain its flow and bypass obstacles, it will likely choose to move from \texttt{Lane~Keeping} to \texttt{Passing} when it detects a hazard. 
Here, the ``simple'' attack fails because the car just drives around the spoofed obstacle. 
To reach the \texttt{Stopped} target, the attacker must adopt a more complex policy (e.g., spoofing a multi-lane blockage) that forces the car to abandon its passing logic follow the path from \texttt{Lane~Keeping} to \texttt{Decelerating}, and eventually to the target \texttt{Stopped} state.

Within CP-ATK\footnote{Code available after publication.} we provide the three above scenarios along with learned optimal attack policies and corresponding utility and Q-value.
We note these are only example configuration files--CP-ATK allows the user to specify the full state machine, action probabilities, and \(\alpha\) values.


\section{Methodology}

The attacker seeks to manipulate the state of the defending CPS by influencing the ML models used.
We model this by saying that at every state \(s \in \mathcal S\) of the CPS, the attacker selects some available action \(a^\dagger \in \mathcal A(s)\) and attempts to fool the models used in that state to lead to that action.

We note the following qualitative observations about this setting.
If the attacker can make the defender reach \(a^\dagger\) in a single step, then this is captured by the standard setting in adversarial learning---the attacker manipulates data so as to trick an ML algorithm.
If the attacker is powerful enough to fully control the defender's actions at every step, then computing the optimal attack is simply the task of finding the shortest path in the defender's state machine.
If, however, \(a^\dagger\) is not reachable in a single step, and the attacker has only limited power, then they seek manipulate the defender's next action so that \(a^\dagger\) is more easily reached in the future.
Because the attacker will assert its (limited) influence as the defender continues through time, the goal is not to compute a single attack action, but instead to compute an {\it attack policy}---a mapping from defender states to attacker actions.

To determine the attack policy, we construct a Markov Decision Process (MDP) by utilizing the attacker's knowledge of the CPS and environment.
We then solve this MDP to determine the attack policy.
We define the MDP \(\tilde {\mathcal M}\) as having the same states and actions as the CPS, but with transition function\footnote{\(T(s, a, s')\) defines the probability of entering state \(s'\) after taking action \(a\) in state \(s\).} \(T: \mathcal S \times \mathcal A \times \mathcal S \rightarrow [0, 1]\) defined as follows:
\begin{align}
\tilde T(s, a^\dagger, s') &= \alpha_{s, a^\dagger}  \mathbbm 1_{[T(s, a^\dagger) = s']} \nonumber\\
&+ (1-\alpha_{s, a^\dagger}) \sum_{a \in \mathcal A(s)} P(a | s) \mathbbm 1_{[T(s, a) = s']}
\end{align}
where \(o\) are the outputs of the models \(\mathcal M (s)\).
Note that the first summand captures the case where the attacker is able to manipulate the action taken.
The second summand utilizes the attacker's knowledge of the defender's behavior (represented by \(P(a | s)\)) and captures the event that the attacker does not affect the selected action (and therefore the CPS continues as it otherwise would).
Solving this MDP yields the attacker's optimal policy. 

To fully define the MDP we must also select the reward function \(R : \mathcal S \rightarrow \mathbb{R}\) and discount factor \(\gamma\).
In solving an MDP, the policy found maximizes the discounted reward\footnote{The expected discounted reward is \(\mathop{\mathbb{E}}_{\bm{s} = [s_0, \ldots s_\infty]}  \sum_i \gamma^i R(s_i).\)} and therefore \(\gamma\) directly captures the ``soft horizon'' discussed above.
One simple reward function is:
\begin{align}
R(s) = \begin{cases} 1~\text{if}~ s = s^\dagger\\
0 ~\text{otherwise}
\end{cases}
\end{align}
As part of this ongoing project, we are investigating alternative reward structures.

\section{Extensions and Discussion}
\label{sec:extensions}

This work marks the beginning of a new long-term project by the authors.
We seek discussion, input, and collaboration from the community.
In what follows we discuss our thoughts on next steps and future directions of CPS security against data manipulation attacks.

In our simplified threat model we assume that the attacker has a fixed probability of attack success for each state.
When presented with input \(\bm x\), the attacker adds vector \(\bm \delta\) to trick the learned model.
When forming the attack policy, \(\bm x\) is unknown and modeling \(P(\texttt{success} ~|~ s)\) makes perfect sense. 
However, when deploying the attack, the attacker may observe \(\bm x\) prior to selecting \(\bm \delta\) and therefore modeling \(P(\texttt{success} ~|~ s, \bm x)\) may be beneficial.

Furthermore, our threat model does not account for the attacker ``missing''.
When the attacker manipulates input data aiming to make the defender take action \(a^\dagger\), we have modeled the attack is either successful (the defender takes action \(a^\dagger\)) or unsuccessful (the defender takes the action \(a\) they would have had no attack occurred).
In practice, though, the attack might yield an action \(a'\) that differs from both \(a\) and \(a^\dagger\). 
How to address this fact (as well as how much it actually changes the optimal strategy for real-world (e.g., vision) models used in CPS) remains as future work.

In addition, we have considered an attacker with a ``soft-deadline'' --- they seek to have the defender research their target state as soon as possible.
If instead they have a hard deadline (``reach this state in the next \(n\) steps''), then the attacker's constructed MDP has a hard horizon and therefore the resulting attack policy is no longer stationary.
This, while mathematically less elegant, can still be solved with standard MDP methods.

Most interesting as future work, though, is the task of defense.
If the defender knows that (some of) its inputs may be affected by an intelligent adversary, how can it make itself more robust?
It can, of course, harden the individual ML methods used, which is a well studied area of research.
A natural question, though, is how one might alter the underlying state machine so as to be more robust while still accomplishing the system's goals.

\section{Related Work}

Research on adversarial ML and the security of CPS has revealed a wide range of vulnerabilities in learning-enabled systems. 
Prior work has studied these vulnerabilities from both the perspective of machine learning models and system-level failures in CPS. 
However, these directions have largely evolved independently, with limited understanding of how adversarial manipulations impact the overall system in a high-level manner. 

\subsection{Adversarial Learning}
Adversarial machine learning has demonstrated that modern deep neural networks are highly vulnerable to carefully crafted perturbations. 
Early work introduced adversarial examples that cause misclassification under minor perturbations imperceptible to the human eye \cite{szegedy_2014_intriguing, goodfellow_2015}. 
These ideas were later extended to stronger optimization-based attacks and robust optimization frameworks \cite{carlini_wagner_2017, madry_2018}.
While these works establish the vulnerability of static perception models, they primarily focus on single-step inference settings. 
In contrast, our work considers sequential decision processes and targets system-level behavior over time.

\subsection{Trustworthy CPS}

Ensuring trustworthiness in CPS has been explored through numerous approaches.
For example, a large body of work focuses on systematically discovering vulnerabilities that undermine CPS trustworthiness.
Fuzzing-based approaches such as AV-Fuzzer \cite{li_2020} and DriveFuzz \cite{kim_2022} generate safety-critical scenarios to expose violations in autonomous driving systems.
More recent efforts extend bug discovery to the semantic and planning layers \cite{wan_2022, huai_2023}.

In addition to internal vulnerabilities, CPS are also susceptible to adversarial manipulation. 
Extensive research has demonstrated \textit{physical-world} attacks against CPS sensors and perception pipelines. 
Prior work has demonstrated the feasibility of physical-world attacks targeting CPS sensing and perception pipelines.
For example, prior research has shown that attackers can manipulate localization through GPS spoofing \cite{shen_2020}, as well as induce misperception via adversarial perturbations to LiDAR and camera inputs \cite{sun_cao_2020, ma_wang_2025, kobayashi_2025}, even in systems that rely on multi-sensor fusion for robustness.
Despite these advances, existing research largely focuses on attacking individual components of CPS, particularly perception modules, without explicitly modeling how these perturbations propagate through the system’s broader decision-making pipeline.
As a result, there remains a limited understanding of how adversaries can systematically manipulate system-level state evolution. 
Our work addresses this gap by constructing adversarial policies that operate directly over the system’s transition dynamics, enabling targeted manipulation of CPS behavior at the level of state trajectories.

\section{Conclusion}
The ultimate goal of this project is to assess and improve the security of AI systems.
We are focused on systems where the composition of learners is explicit and captured via an underlying state machine.
Cyber-Physical Systems (CPS) often fit into this category and are our main motivation.

In this preliminary and on-going work we consider an attacker aiming to manipulate the state of the defender.
We model an attacker with limited but continued influence as the defender acts in time.
As such, the goal of the attacker is not to determine a single fire-and-forget attack, but rather an {\it attack policy} defining what action to take in every state.
For simplified settings we reduced the problem of computing an optimal attack policy to solving an induced MDP.

We hope this paper, paired with CP-ATK, serves as a springboard for discussions with the community about the security of cyber-phsyical systems and compositional learning more broadly.

\bibliography{refs}
\bibliographystyle{icml2026}

\end{document}